\def\year{2019}\relax
\documentclass[letterpaper]{article} 
\usepackage{aaai19}  
\usepackage{times}  
\usepackage{helvet}  
\usepackage{courier}  
\usepackage{url}  
\usepackage{graphicx}  
\usepackage{comment}
\usepackage{subfig}
\usepackage{csquotes} 
\usepackage{devanagari} 
\usepackage{footmisc}
\usepackage{graphicx}
\makeatletter
\def\endthebibliography{%
  \def\@noitemerr{\@latex@warning{Empty `thebibliography' environment}}%
  \endlist
}
\makeatother

\begin{document}
%
\title{Mind Your Language: Abuse and Offense Detection for Code-Switched Languages}
\author{
Raghav Kapoor\\ MIDAS Lab, NSIT-Delhi\\raghavk.co@nsit.net.in
\And Yaman Kumar \\ Adobe Systems \\ykumar@adobe.com
\And Kshitij Rajput \\ MIDAS Lab, NSIT-Delhi\\kshitijr.co@nsit.net.in
\AND Rajiv Ratn Shah \\IIIT, Delhi\\rajivratn@iiitd.ac.in
\And Ponnurangam Kumaraguru \\IIIT, Delhi \\pk@iiitd.ac.in
\And Roger Zimmermann \\NUS, Singapore\\rogerz@comp.nus.edu.sg
}
\maketitle
\begin{abstract}
In multilingual societies like the Indian subcontinent, use of code-switched languages is much popular and convenient for the users. In this paper, we study offense and abuse detection in the code-switched pair of Hindi and English (i.e. Hinglish), the pair that is the most spoken. The task is made difficult due to non-fixed grammar, vocabulary, semantics and spellings of Hinglish language. We apply transfer learning and make a LSTM based model for hate speech classification. This model surpasses the performance shown by the current best models to establish itself as the state-of-the-art in the unexplored domain of Hinglish offensive text classification. We also release our model and the embeddings trained for research purposes.
\end{abstract}

\section{Introduction}
\label{introduction}
With the penetration of internet among masses, the content being posted on social media channels has uptaken. Specifically, in the Indian subcontinent, number of Internet users has crossed 500 mi\footnote{\url{https://bit.ly/2MCXz2Q
%https://timesofindia.indiatimes.com/business/india-business/number-indian-internet-users-will-reach-500-million-by-june-2018-iamai-says/articleshow/62998642.cms
}}, and is rising rapidly due to inexpensive data\footnote{\url{https://bit.ly/2NUjOGh
%https://www.huffingtonpost.in/2017/06/01/reliance-jio-is-driving-indian-internet-growth-says-the-mary-me_a_22120777/
}}. With this rise, comes the problem of hate speech, offensive and abusive posts on social media. Although there are many previous works which deal with Hindi and English hate speech (the top two languages in India), but very few on the code-switched version (Hinglish) of the two \cite{mathur2018detecting}. This is partially due to the following reasons: (i) Hinglish consists of no-fixed grammar and vocabulary. It derives a part of its semantics from Devnagari and another part from the Roman script. (ii) Hinglish speech and written text consists of a concoction of words spoken in Hindi as well as English, but written in the Roman script. This makes the spellings variable and dependent on the writer of the text.
Hence code-switched languages present tough challenges in terms of parsing and getting the meaning out of the text. For instance, the sentence, \emph{``Modiji foreign yatra par hai''}, is in the Hinglish language. Somewhat correct translation of this would be, \emph{``Mr. Modi is on a foriegn tour''}. However, even this translation has some flaws due to no direct translation available for the word \emph{ji}, which is used to show respect. Verbatim translation would lead to \emph{``Mr. Modi foreign tour on is''}. Moreover, the word \emph{yatra} here, can have phonetic variations, which would result in multiple spellings of the word as \emph{yatra, yaatra, yaatraa}, etc. Also, the problem of hate speech has been rising in India, and according to the policies of the government and the various social networks, one is not allowed to misuse his right to speech to abuse some other community or religion. Due to the various difficulties associated with the Hinglish language, it is challenging to automatically detect and ban such kind of speech.

Thus, with this in mind, we build a transfer learning based model for the code-switched language Hinglish, which outperforms the baseline model of \cite{mathur2018detecting}. We also release the embeddings and the model trained. 
\begin{table}[]
    \centering
    \begin{tabular}{|c|c|c|c|}
        \hline
    \textbf{Hinglish} & \textbf{English} & \textbf{Hinglish} & \textbf{English} \\ \hline 
        acha & good & gunda & thug \\
        s**la & blo*dy & ra*di & h*oker \\
         \hline
    \end{tabular}
    \caption{Examples of word-pairs in Hinglish-English dictionary}
    \label{tab:examples_hinglish_english}
\end{table}

\begin{table*}[]
    \centering
    \begin{tabular}{|c|c|c|}
    \hline
        \textbf{Category} & \textbf{Tweet} & \textbf{Translation} \\ \hline 
        Benign & sache sapooto aap ka balidan hamesha yaad rahega & True sons, your sacrifice would be remembered.\\
        Hate Inducing & Bik gya Porkistan & Porkistan (Derogatory term for Pakistan) has been sold\\
        Abusive & Kis m*darch*d ki he giri hui harkt & Which m*therf*cker has done this \\
         \hline
    \end{tabular}
    \caption{Examples of tweets in the dataset and their translations}
    \label{tab:examples_tweets}
\end{table*}
\section{Methodology}
Our methodology primarily consists of these steps: Pre-processing of the dataset, training of word embeddings, training of the classifier model and then using that on HEOT dataset.
\subsection{Pre-Processing}
In this work, we use the datasets released by \cite{davidson2017automated} and HEOT dataset provided by \cite{mathur2018detecting}. The datasets obtained pass through these steps of processing: (i) Removal of punctuatios, stopwords, URLs, numbers, emoticons, \emph{etc.} This was then followed by transliteration using the Xlit-Crowd conversion dictionary \footnote{\url{https://github.com/chsasank/indic-transliteration}} and translation of each word to English using Hindi to English dictionary\footnote{\url{http://www.cfilt.iitb.ac.in/~hdict/webinterface_user/}}. To deal with the spelling variations, we manually added some common variations of popular Hinglish words. Final dictionary comprised of 7200 word pairs. Additionally, to deal with profane words, which are not present in Xlit-Crowd, we had to make a profanity dictionary (with 209 profane words) as well. Table \ref{tab:examples_hinglish_english} gives some examples from the dictionary.

\subsection{Training Word Embeddings}
We tried Glove \cite{pennington2014glove} and Twitter word2vec \cite{godin2015multimedia} code for training embeddings for the processed tweets. The embeddings were trained on both the datasets provided by \cite{davidson2017automated} and HEOT. These embeddings help to learn distributed representations of tweets. After experimentation, we kept the size of embeddings fixed to 100.
\begin{table}[]
    \centering
    \begin{tabular}{|c|c|}
        \hline
    \textbf{Model} &  \textbf{Accuracy}\\ \hline 
        Davidson et al. & 0.57\\
        Our Model with embeddings trained on Glove & \textbf{0.87} \\
        Our Model with embeddings trained on Word2Vec & 0.82\\
        Our Model with pre-trained Word2Vec embeddings  & 0.59 \\
        Mathur et al & 0.83 \\
         \hline
    \end{tabular}
    \caption{Comparison of accuracy scores on HEOT dataset}
    \label{tab:results_heot}
\end{table}

\begin{table}[]
    \centering
    \begin{tabular}{|c|c|}
        \hline
    \textbf{Model} &  \textbf{Accuracy}\\ \hline 
        Davidson et al. & \textbf{0.90}\\
Our Model with embeddings trained on Glove & 0.89 \\
Our Model with embeddings trained on Word2Vec & 0.86 \\
Mathur et al. & 0.75 \\
         \hline
    \end{tabular}
    \caption{Comparison of accuracy scores on \cite{davidson2017automated} dataset}
    \label{tab:results_davidson}
\end{table}
\subsection{Classifier Model}
\begin{figure}
\centering
\includegraphics[scale = 0.28]{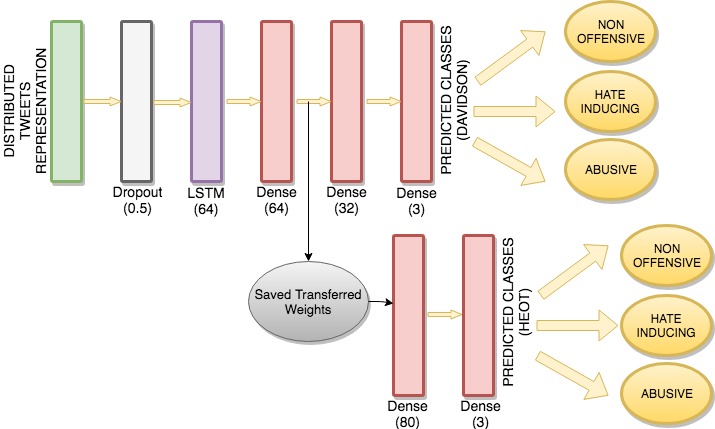}
\caption{LSTM based model for tweet classification}
\label{ml_model}
\end{figure}

Both the HEOT and \cite{davidson2017automated} datasets contain tweets which are annotated in three categories: offensive, abusive and none (or benign). Some examples from the dataset are shown in Table \ref{tab:examples_tweets}. We use a LSTM based classifier model for training our model to classify these tweets into these three categories. An overview of the model is given in the Figure \ref{ml_model}. The model consists of one layer of LSTM followed by three dense layers. The LSTM layer uses a dropout value of 0.2. Categorical crossentropy loss was used for the last layer due to the presence of multiple classes. We use Adam optimizer along with L2 regularisation to prevent overfitting. As indicated by the Figure \ref{ml_model}, the model was initially trained on the dataset provided by \cite{davidson2017automated}, and then re-trained on the HEOT dataset so as to benefit from the transfer of learned features in the last stage. The model hyperparameters were experimentally selected by trying out a large number of combinations through grid search.

\section{Results}
Table \ref{tab:results_heot} shows the performance of our model (after getting trained on \cite{davidson2017automated}) with two types of embeddings in comparison to the models by \cite{mathur2018detecting} and \cite{davidson2017automated} on the HEOT dataset averaged over three runs. We also compare results on pre-trained embeddings. As shown in the table, our model when given Glove embeddings performs better than all other models. For comparison purposes, in Table \ref{tab:results_davidson} we have also evaluated our results on the dataset by \cite{davidson2017automated}.

\section{Conclusion}
In this paper, we presented a pipeline which given Hinglish text can classify it into three categories: offensive, abusive and benign. This LSTM based model performs better than the other systems present. We also release the code, the dictionary made and the embeddings trained in the process. We believe this model would be useful in hate speech detection tasks for code-switched languages.

\fontsize{9.0pt}{10.0pt}
\selectfont
\bibliographystyle{aaai}
\bibliography{bibliography}
\end{document}